# Research on Chinese News Summary: Research on Information Extraction of LCSTS Dataset Based on an Improved BERTSum-LSTM Model

*Yiming Chen, College of Information Science and Technology, Donghua University, Shanghai, 201620, China, 210995109@mail.dhu.edu.cn*

*Haobin Chen, College of Information Science and Technology, Donghua University, Shanghai, 201620, China, hb_cccc@163.com*

*Simin Liu, College of Physics, Donghua University, Songjiang, Shanghai, 201620, China, 220120114@mail.dhu.edu.cn*

*Yunyun Liu, Glorious Sun School of Business & Management, Donghua University, Shanghai, 200051, China, cemoyang@163.com*

*Fanhao Zhou, College of Medical Technology, Shanghai University of Medicine & Health Sciences, Shanghai, 201318, China, b21050103018@stu.sumhs.edu.cn*

*Bing Wei, College of Information Sciences and Technology, Donghua University, Shanghai, 201620, China, bingwei@dhu.edu.cn*

***Abstract:*** *With the continuous advancement of artificial intelligence, natural language processing technology has become widely utilized in various fields. At the same time, there are many challenges in creating Chinese news summaries. First of all, the semantics of Chinese news is complex, and the amount of information is enormous. Extracting critical information from Chinese news presents a significant challenge. Second, the news summary should be concise and clear, focusing on the main content and avoiding redundancy. In addition, the particularity of the Chinese language, such as polysemy, word segmentation, etc., makes it challenging to generate Chinese news summaries. Based on the above, this paper studies the information extraction method of the LCSTS dataset based on an improved BERTSum-LSTM model. We improve the BERTSum-LSTM model to make it perform better in generating Chinese news summaries. The experimental results show that the proposed method has a good effect on creating news summaries, which is of great importance to the construction of news summaries.*

***Keywords:*** *News summary; Artificial intelligence; Natural language processing; Message requirement; Core factor extraction*

## 1. Introduction

News summary is one of the primary responsibilities in natural language processing and is also part of information retrieval and text understanding. It can be divided into extracted and generated summaries comprising statistical machine learning and deep learning. Researchers commissioned computers to generate automated news summaries to improve the efficiency of information processing. Since the popularity of the Internet, news summaries have become the key to information overload, and the summary quality is an evaluation index. Unlike traditional information retrieval, news summaries emphasize timeliness, accuracy, and readability [1]. Therefore, we propose the topic of automated news summaries, and artificial intelligence creates possibilities for improving news summaries.

News summary originates from computer science, with information processing as the core. Its algorithm shows a deep understanding of language and text and is also a tool for information dissemination. From a technical perspective, news summary highlights simplicity and comprehensiveness, modernizing information processing through algorithms and linguistics. And it is on the theoretical level. Today, news summaries have undergone a unique transformation from laboratory to practical application. The comprehensive promotion of artificial intelligence refined the news summary technology and embodied technological progress. In addition, it affects how information is disseminated and profoundly impacts the news industry. Hence, in discussing news summaries, we must be forward-looking and strategic. In the new era, we propose the issue of high-quality development of news summaries.

In conclusion, a news summary is a critical condition and guarantee of the efficient dissemination of information. From a technological development standpoint, news summaries have made significant progress. However, technical personnel have yet to find an effective way to extract high-quality abstracts, and they are making efforts. Therefore, a news summary requires continuous exploration and innovation, which is essential to meeting people's growing information needs and is a vital part of advancing information technology [2].

Based on the analysis above, this paper proposes a news summary model that integrates deep learning and natural language processing technology to enhance the quality and efficiency of news summaries. We solve the critical problems in news summary generation by constructing a theoretical analysis framework and simulation experiments. The main contents include the description and simulation analysis of the Chinese news summary model, which effectively deals with the risk of information overload and has both theoretical and practical significance.

## 2. A Review of Research on News Summary Generation

From a technical point of view, the generation of news summaries is the essential link of natural language processing and the embodiment of information retrieval and text understanding. Therefore, the main logic of news summary research is automatic summary generation. Automatic summary is the leading research object of news summary, and it is also an innovative subject in information processing [3]. Currently, the research of news summaries strengthens quality control from the perspective of algorithms, and there are three primary forms: the first form is an extracted summary. The algorithm realizes effective communication between text understanding and information extraction. The second form is the generative summary [4]. By establishing a language model and generation standard and making the generation standard public to users, the standardization control is realized. The third is the internal process of reengineering abstract generation. In recent years, advancements in deep learning and natural language processing technology have led to improvements in the quality of summaries and the efficiency of information processing through intelligent means. However, in the face of diverse user needs, the intelligence level of news summaries needs to be further improved.

## 3. Main Application Models

*3.1 LSTM Model*

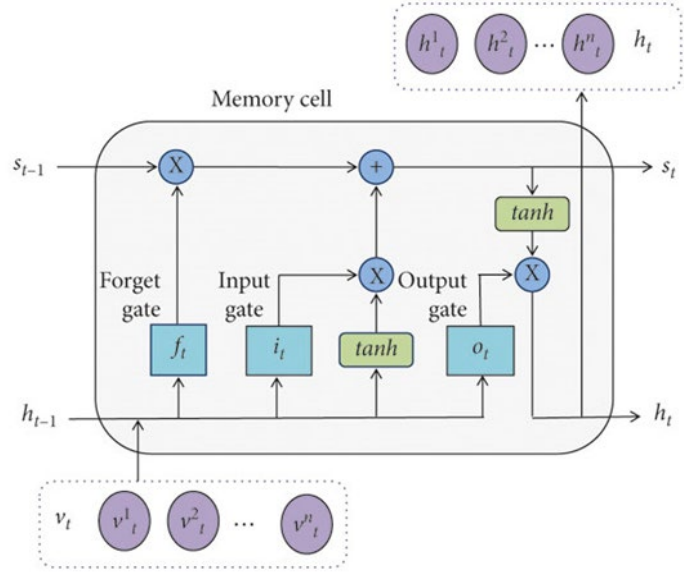

Figure 1 Schematic diagram of LSTM model structure

LSTM (Long Short-Term Memory) is an improved RNN (Recurrent Neural Network) that processes sequence data and understands context information. The introduction of LSTM solves the gradient disappearance and gradient explosion of RNN when dealing with long sequence data and performs well in long-term dependent tasks [5].

The extractive model creates a summary by selecting the most important sentences in the article. In the extractive LSTM summary model, the text is encoded into the LSTM, and the importance score of each sentence is obtained after entering the linear layer. Finally, it outputs the most important one or more sentences by sorting.

We use the following formulas to represent the basic structure of the LSTM unit, in order to describe this process in more detail:

$$f_t = \sigma(W_f \cdot [h_{t-1}, x_t] + b_f)$$

$$i_t = \sigma(W_i \cdot [h_{t-1}, x_t] + b_i)$$

$$\tilde{C}_t = \tanh(W_C \cdot [h_{t-1}, x_t] + b_C) \quad (1)$$

$$C_t = f_t \circ C_{t-1} + i_t \circ \tilde{C}_t$$

$$o_t = \sigma(W_o \cdot [h_{t-1}, x_t] + b_o)$$

$$h_t = o_t \circ \tanh(C_t) \quad (2)$$

In this formula:

- $f_t$: It stands for forget gate and determines which information should be discarded from the cell state.

- $i_t$: It stands for input gate and determines what new information will be added to the cell state.

- $\tilde{C}_t$: It represents the new candidate cell state and contains the new information currently entered.

- $C_t$: It represents an updated cellular state and is a combination of forget and input gate results.

- $o_t$: It represents the output gate and determines the information that the next hidden state should contain.

- $h_t$: The next hidden state is calculated based on the cell state and output gate.

- $W_f, W_i, W_C, W_o$: The weight matrix is used to calculate the gating function and candidate cell state.

- $b_f, b_i, b_C, b_o$: The bias term is used to adjust the output of the gating function and the candidate cell state.

- $\sigma$: A sigmoid function compresses the input between 0 and 1 for gating.

- $\tanh$: The hyperbolic tangent function generates the candidate cell state and hidden state.

- $\circ$: It represents element-by-element multiplication, which is used to implement the gating function.

This formula explains how the LSTM unit maintains and updates the cell state through a gating mechanism to generate the hidden state, which is the crucial mechanism of LSTM when processing sequence data.

*3.2 BERTSum Model*

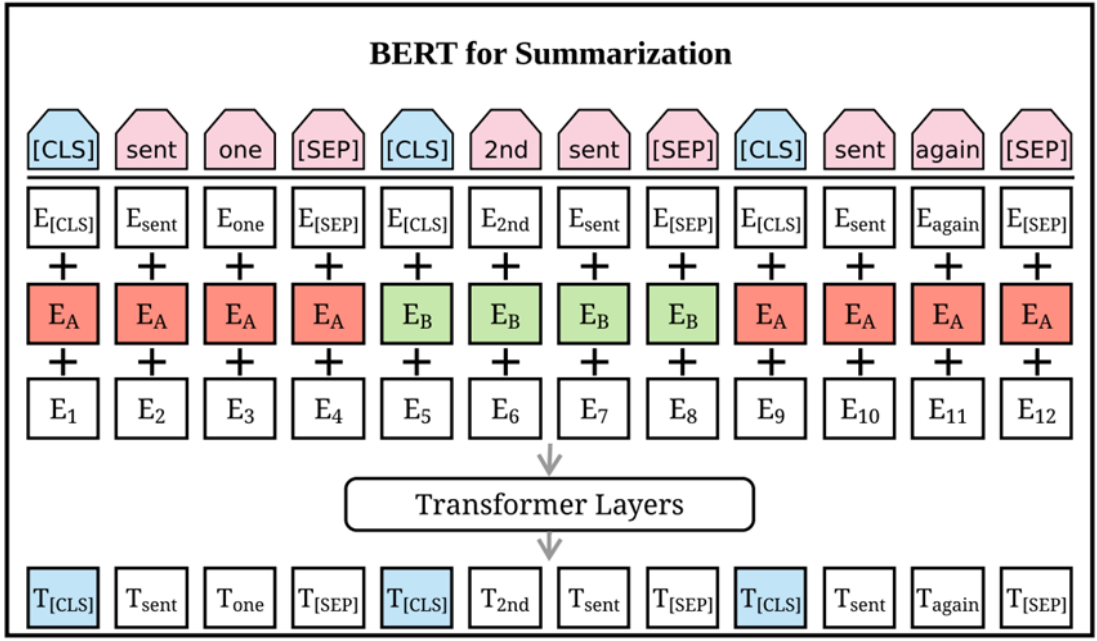

Figure 2 BERTSum model structure

The extractive BERTSum works similarly to the extractive LSTM. It uses a deep network with a BERT-like structure to transform the paragraph into a sentence embedding. These sentences are embedded into a linear layer for scoring and sorting. This process helps predict which sentences should be selected as a summary [6].

The BERTSum model is an extractive summary model based on the pre-trained language model BERT. It uses the BERT model to encode the text to obtain the deep semantic representation of each sentence and calculates the importance score of sentences through an additional linear layer. Finally, the most important sentences are selected according to the score to generate the summary.

We use the following formulas to explain and describe the summary generation process of the BERTSum model in more detail:

$$s_i = \text{BERT}(x_i)$$

$$h_i = \tanh(W_h s_i + b_h) \quad (3)$$

$$p_i = \text{softmax}(W_p h_i + b_p)$$

- $h_i$: t represents the output of $s_i$ after passing it through a linear layer and performing a tanh activation function, which is used to calculate the importance of each sentence.

- $W_p$ and $b_p$ are the weights and bias of another linear layer, used to map h_ihi to the probability space.

- $W_h$ and $b_h$ represent the weight and bias of the linear layer.

- $p_i$ denotes the probability that the i-th sentence will be selected as the summary after being processed by the softmax function [7].

- $W_p$ and $b_p$ represent the weight and bias of another linear layer, respectively, which maps $h_i$ to the probability space.

These formulas show how the BERTSum model uses the BERT model to encode the text, then calculates the importance value of each sentence through an additional linear layer, and finally applies the softmax function to determine the probability of each sentence chosen as a summary.

### 3.3 BERTSum-LSTM Model

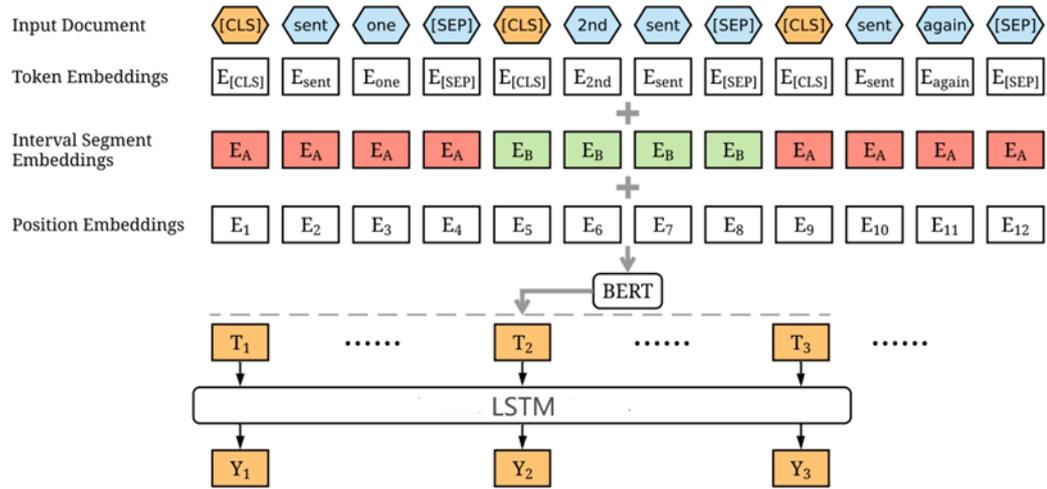

Figure 3 BERTSum-LSTM model structure

The BERTSum-LSTM model we proposed combines the advantages of BERTSum and LSTM. BERTSum-encoded sentence embeddings are fed into LSTM to extract context information. In this way, the model can not only capture the global context information but also extract the local features to perform well in the text summarization task.

The BERTSum-LSTM model is an extractive summarization model that combines BERT and LSTM. It encodes the text with the BERT model to obtain the deep semantic representation of each sentence. Next, the LSTM model is used to capture the sequence dependencies in the text. Finally, a linear layer is used to calculate the importance score of each sentence, and the most vital sentence is selected according to the score to generate a summary.

We use the following formulas to represent the summary generation process of the BERTSum-LSTM model and show the details.

$$s_i = \text{BERT}(x_i)$$

$$h_i = \text{LSTM}(s_i) \tag{4}$$

$$p_i = \text{softmax}(W_p h_i + b_p)$$

In the formulas:

- $s_i$ represents the deep semantic representation of the i-th sentence after being encoded by the BERT model.

- $BERT(x_i)$ means that the i-th sentence $x_i$ is input into the BERT model to obtain its corresponding deep semantic representation.

- $h_i$ represents the hidden state obtained by inputting $s_i$ into the LSTM model to capture sequence dependencies.

- $\text{LSTM}(s_i)$ represents the input of $s_i$ into the LSTM model to obtain the hidden state.

- $p_i$ is the probability that the i-th sentence is selected as the summary after being processed by the softmax function.

- $W_p$ and $b_p$ are the weights and biases of the linear layer, respectively, which are used to map $h_i$ to the probability space.

These formulas show how the BERTSum-LSTM model uses the BERT model to encode the text. The LSTM model is used to capture the sequence dependencies in the text. In addition, it calculates the importance value of each sentence through a linear layer and finally applies the softmax function to determine the probability of each sentence chosen as a summary.

**4. Data Description**

The research topic proposed points out that the main direction of the research is the generation of Chinese news summaries, especially by improving the BERTSum-LSTM model to optimize the process. Moreover, this study is conducted on the LCSTS (Large Scale Chinese Short Text Summarization) dataset. Here is an overview of this section:

This section elaborates on the key information such as the characteristics, structure, scale, and preprocessing methods of the LCSTS dataset used in the study.

Here are some key points that will be covered:

(1) Introduction to the dataset

(2) LCSTS overview: This includes the origin of the LCSTS dataset, the purpose of its creation, its importance in the field of natural language processing, and why it is suitable for the study of Chinese news summaries.

(3) Dataset characteristics: This section explains the uniqueness of the dataset, such as text diversity (covering different news fields), the manual generation of summaries to ensure quality, and the characteristics of short texts.

(4) Dataset size: It shows the number of samples. Here is a total of news article pairs and corresponding summaries in the dataset, indicating the dataset's size and suitability for model training and testing.

(5) Category distribution: If the news in the dataset is categorized by topic or type, the dataset will provide a breakdown of the number of samples in each category and analyze the impact of category balance on model training.

(6) Data preprocessing: In text cleaning, the preprocessing steps of the original text are introduced. These steps include removing irrelevant symbols, normalizing digits, segmenting words, and removing stop words to match the model input format. Serialization and coding explain how to convert text into a sequence of numbers the model can understand, including word embedding or the tokenization process of pre-training the model using BERT. It addresses the data partitioning process, detailing how the dataset is split into training, validation, and test sets. It covers proportional allocation and randomization strategies to ensure fair and generalized model evaluation.

(7) Challenges and responses: It analyzes the noise problem, discusses the possible noise in the data set (such as mislabeling and inconsistency), and measures to reduce these effects. For the long-tail phenomenon, if it exists, we will analyze the long-tail distribution in the data set and its potential impact on the model training, as well as how to deal with it through methods such as oversampling, undersampling, or weight adjustment.

The above content provides readers with critical information to gain a deeper understanding of the research fundamentals, provides a complete picture of the dataset, and explains the challenges faced by the research team in preparing the data and associated solutions. It lays a solid foundation for subsequent model building and analysis of experimental results.

*4.1 LCSTS Dataset*

The LCSTS dataset is a large-scale, high-quality Chinese short-text summary dataset. The dataset contains more than 2 million Chinese short-text data written by real people and abstracts given by the authors. As the dataset includes abstract texts, it can not be used to train the abstract summary model. To address this, we pre-process the dataset and apply the greedy algorithm to select sentences that maximize the ROUGE score between the answers and the selected sentences. This way, we get the summary sentence number list suitable for training the extractive summarization model.

Table 1 The performance comparison of the models on the LCSTS dataset

| Model | ROUGE-1 | ROUGE-2 | ROUGE-L |
|---|---|---|---|
| LSTM | 57.73 | 45.55 | 49.57 |
| BERTSum | 55.00 | 42.33 | 46.85 |
| BERTSum-LSTM | 62.29 | 50.64 | 51.49 |

The BERTSum-LSTM model shows superior performance compared to the other two models on the test set, achieving ROUGE-1 of 62.29%, ROUGE-2 of 50.64%, and ROUGE-L of 51.45%.

*4.2 Quantification and Indicators*

ROUGE (Recall-Oriented Understudy for Gisting Evaluation) is a metric for evaluating the quality of automatic text summarization and machine translation. ROUGE assesses text quality by comparing the overlap between machine-generated text and reference text. Here are the commonly used ROUGE scores:

ROUGE-N: Matching based on n-gram. ROUGE-1 (unigram matching) and ROUGE-2 (bigram matching) are common. It computes the number of n-gram overlaps between the generated text and the reference text.

ROUGE-L: It is based on the Longest Common Subsequence (LCS). It measures the longest common subsequence between the generated text and the reference text to assess the overall text similarity.

**5. Simulation and Analysis**

Simulation technology plays a "core role" as a bridge between theory and practice when simulating the behavior of complex systems. In engineering design and optimization mechanisms, simulation is considered a standard and effective verification method, which plays an important role in predicting system performance and reducing the cost of physical prototype testing. Simulation is not only a technical concept but also related to methodology. Therefore, the strategy of "simulation first" has become the key to improving research efficiency. Overall, the practical deduction of simulation is a problem-solving method gradually formed based on computer simulation, which involves much trial and error. From establishing the initial model to fine-tuning the parameters, the simulation process is closely linked to improving prediction accuracy and efficiency. Simulation technology aims to closely represent the real world to fulfill the requirements of high-precision simulation. However, while pursuing high efficiency, a dilemma arises: a gap exists between simulation and actual operation. Generally, model accuracy and real-time response speed of simulation technology need improvement. Moreover, the algorithm's robustness and generalization ability need further improvement, which are crucial for the future development of simulation technology.

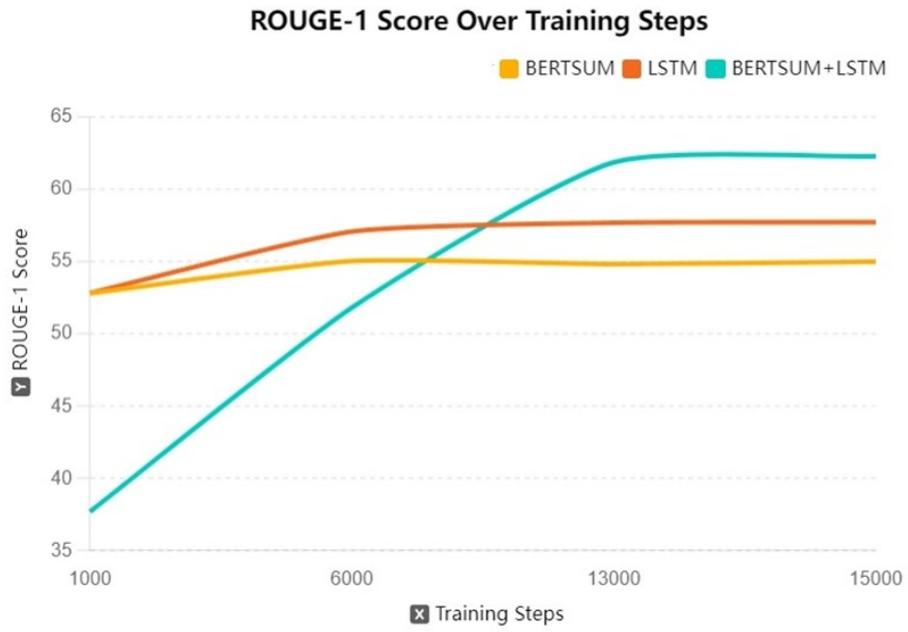

Figure 4 Analysis of ROUGE-L simulation results

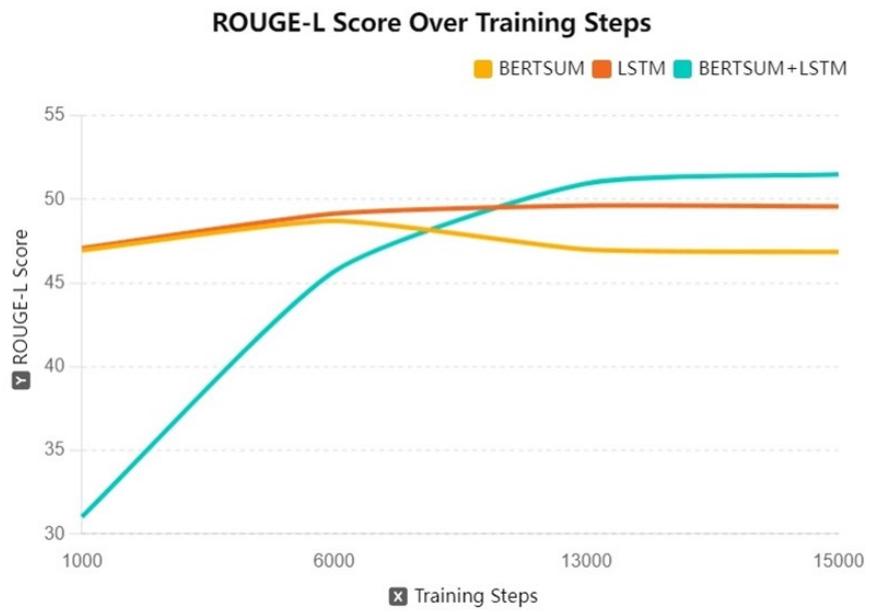

Figure 5 Simulation data analysis of ROUGE-L

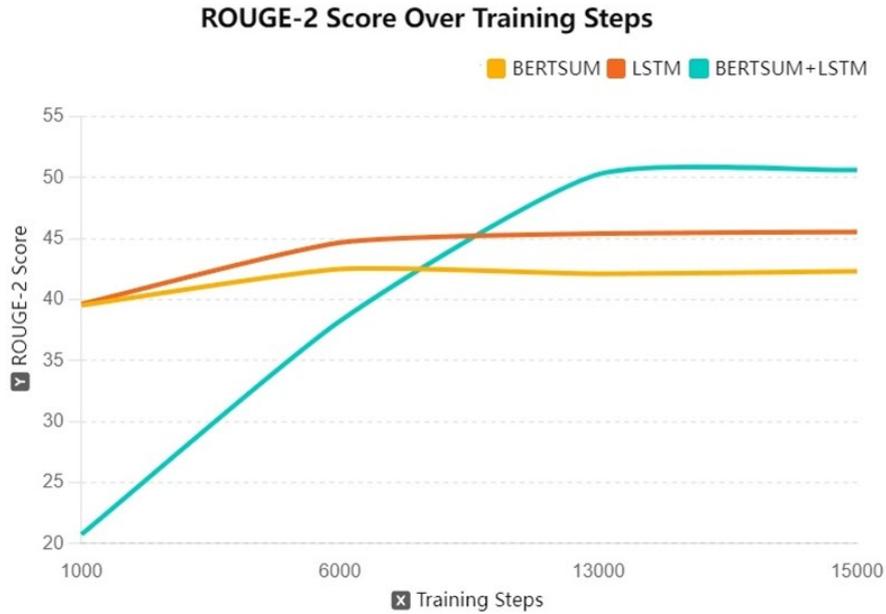

Figure 6 Simulation data analysis of ROUGE-2

We conducted a cause analysis.

(1). The BERTSum layer processes the text sequence, extracts sentence vector information, and extracts features between sentences through the LSTM layer. Multi-layer feature extraction enables the model to gain a more comprehensive understanding of the semantic information in the text, leading to more accurate evaluation and sorting of sentences.

(2). The BERTSum model has a deeper network structure. Paired with LSTM, it creates a deeper network structure. This allows the BERTSum-LSTM structure to more effectively capture the intricate relationships and features in the text, thereby enhancing model performance.

**6. Conclusion**

Chinese news summary has reached a new stage in applying the deep learning model, which puts forward new challenges and requirements for automatic content understanding and generation technology. A news summary is a representation of "efficient" information concentration and a vital tool to enhance information dissemination efficiency. In addition, it is conducive to accurate knowledge transfer and maintains the diversity of information. It reflects the requirements of artificial intelligence technology for the media industry. Due to the continuous progress of natural language processing technology, in the research based on the improved BERTSum-LSTM model, we have built a theoretical analysis framework and practical optimization mechanism for LCSTS data sets. In recent years, the integrated application of modern information technologies, such as the BERT model and LSTM, has advanced innovation in the field of natural language processing. Deep learning has facilitated the completion of text summarization tasks and improved abstract content's accuracy and scientific quality. The improved BERTSum-LSTM model research aligns with the information processing in the intelligent era. It offers a new approach for automating and enhancing the quality of Chinese news summaries. In a word, the sustainable improvement and development of this model will better meet the demand of rapid consumption in the information age and promote the transformation of the media industry to be more intelligent and efficient.

We compare and study many methods to solve the problem of extractive summarization of Chinese news and propose a deep neural network architecture, which has achieved significant improvement on the Chinese news summary dataset. Our BERTSum-LSTM model performs better than the BERTSum and LSTM models on the LCSTS dataset.

The improvements that need to be made in the future are as follows. First, at present, for generative summary, the model can only complete the task by selecting the most important sentences in the paragraph. Introducing large models for generative summary or simply drawing on their knowledge may help improve the model's ability. Second, our proposed model works well in domain-specific text summarizing, especially news texts. Researchers may try to apply this model to more advanced tasks，such as summarizing conferences in the future. Third, a limitation of this experiment is that it only summarizes Chinese texts. Adapting the model to multiple languages is an important research direction. Fourth, when it comes to speech summarization, for some languages that lack written text, processing their speech is often the only option. Adapting our model to speech presents a more challenging task.

## References

*[1]. Wan X, Zhang J. CTSUM: extracting more certain summaries for news articles[C]//Proceedings of the 37th international ACM SIGIR conference on Research & development in information retrieval. 2014: 787-796.*
*[2]. Soni M, Wade V. Comparing abstractive summaries generated by chatgpt to real summaries through blinded reviewers and text classification algorithms[J]. arXiv preprint arXiv:2303.17650, 2023.*
*[3]. Gambhir M, Gupta V. Recent automatic text summarization techniques: a survey[J]. Artificial Intelligence Review, 2017, 47(1): 1-66.*
*[4]. Ding J, Li Y, Ni H, et al. Generative text summary based on enhanced semantic attention and gain-benefit gate[J]. IEEE Access, 2020, 8: 92659-92668.*
*[5]. Sherstinsky A. Fundamentals of recurrent neural network (RNN) and long short-term memory (LSTM) network[J]. Physica D: Nonlinear Phenomena, 2020, 404: 132306.*
*[6]. Nunna J L D, Hanuman Turaga V K, Chebrolu S. Extractive and abstractive text summarization model fine-tuned based on bertsum and bio-bert on covid-19 open research articles[C]//International Conference on Machine Learning and Big Data Analytics. Cham: Springer International Publishing, 2022: 213-223.*
*[7]. Zhang Y, Zhang Y, Peng L, et al. Base-2 softmax function: Suitability for training and efficient hardware implementation[J]. IEEE Transactions on Circuits and Systems I: Regular Papers, 2022, 69(9): 3605-3618.*